\documentclass[sigconf, nonacm]{acmart}
\AtBeginDocument{%
  \providecommand\BibTeX{{%
    \normalfont B\kern-0.5em{\scshape i\kern-0.25em b}\kern-0.8em\TeX}}}

\setcopyright{none}
\usepackage{url}

\usepackage{graphicx}

\usepackage[utf8]{inputenc}

\usepackage{hyperxmp}
\usepackage{fancyhdr}

\usepackage{lipsum}
\usepackage{multicol}

\usepackage{caption}
\usepackage{subcaption}

\begin{document}

\title{A Comparison of Supervised and Unsupervised Deep Learning Methods for Anomaly Detection in Images}

\author{Vincent Wilmet}
\email{vincent.wilmet@student-cs.fr}
\affiliation{
  \institution{CentraleSupélec}
  \city{Gif-sur-Yvette}
  \country{France}
}
\author{Sauraj Verma}
\email{saureyj.verma@student-cs.fr}
\affiliation{
  \institution{CentraleSupélec}
  \city{Gif-sur-Yvette}
  \country{France}
}

\author{Tabea Redl}
\email{tabea.redl@student-cs.fr}
\affiliation{
  \institution{CentraleSupélec}
  \city{Gif-sur-Yvette}
  \country{France}
}

\author{Håkon Sandaker}
\email{hakon.sandaker@student-cs.fr}
\affiliation{
  \institution{CentraleSupélec}
  \city{Gif-sur-Yvette}
  \country{France}
}

\author{Zhenning Li}
\email{zhenning.li@student-cs.fr}
\affiliation{
  \institution{CentraleSupélec}
  \city{Gif-sur-Yvette}
  \country{France}
}


\begin{abstract}
Anomaly detection in images plays a significant role for many applications across all industries, such as disease diagnosis in healthcare or quality assurance in manufacturing. Manual inspection of images, when extended over a monotonously repetitive period of time is very time consuming and can lead to anomalies being overlooked.Artificial neural networks have proven themselves very successful on simple, repetitive tasks, in some cases even outperforming humans. Therefore, in this paper we investigate different methods of deep learning, including supervised and unsupervised learning, for anomaly detection applied to a quality assurance use case. We utilize the MVTec anomaly dataset and develop three different models, a CNN for supervised anomaly detection, KD-CAE for autoencoder anomaly detection, NI-CAE for noise induced anomaly detection and a DCGAN for generating reconstructed images. By experiments, we found that KD-CAE performs better on the anomaly datasets compared to CNN and NI-CAE, with NI-CAE performing the best on the Transistor dataset. We also implemented a DCGAN for the creation of new training data but due to computational limitation and lack of extrapolating the mechanics of AnoGAN, we restricted ourselves just to the generation of GAN based images. We conclude that unsupervised methods are more powerful for anomaly detection in images, especially in a setting where only a small amount of anomalous data is available, or the data is unlabeled.
\end{abstract}


\keywords{Anomaly detection, Computer vision, Convolutional Neural Networks, Convolutional Autoencoders, Generative Adversarial Networks}


\maketitle

\paragraph{Code}
Link to the git repository containing all code used for this paper: \href{https://github.com/vincentwi/Anomaly-Detection}{https://github.com/vincentwi/Anomaly-Detection}

\section{Introduction}
Anomaly, also synonymous with the term "outlier", is a deviant observation that is different from the distribution of the provided data. Often in the arena of machine learning and statistical modeling, anomalies are some of the key observations that hold the power to shift the balance of interpretation and modeling of complex dynamical systems, whether it be predicting some classification score or making statistical inferences about the population parameters.
\par 
Often for us humans, detecting anomalies and deviant patterns amongst data is fairly easy since our years of evolutionary training gives us the advantage to distinguish signal and noise easily, but this is often not the case for intelligent systems as they still struggle to effectively identify various forms of anomalies and unknown patterns within the information they are fed. This often leads us to the problem of anomaly detection, where the central goal is to detect these deviant observations and to be able to deduce the properties that make them an outlier. Anomaly detection has become one of the most insightful problems in the modern-day world since various industries and business organizations strive to develop systems that are not only robust to anomalies in incoming data but also are able to detect them appropriately, classify them as an outlier and let machine learning experts know about their existence so that they can be tackled appropriately.
\par 
Such anomalous events pose a challenge not only because of their complexity in nature but also due to their infrequent appearance and limited evidence of causality; for most machine learning and expert systems, detection and classification of these anomalies is often a challenge due to the lack of data that sufficiently allows them to generalize their properties and accurately label them as an outlier. The limited occurrences of such events are what cause problems in the detection phase and lead to the imbalanced data classification problem. In cases of class imbalance, it is often not the norm to focus on model accuracy since it has been observed that model accuracy induces bias towards the majority class, whereas the minority class which is the target of interest is left underrepresented \cite{weiss2000learning}.
\par 
One of the most prominent and promising methodologies for anomaly detection lies in deep learning, a sub-field of machine learning that focuses on using deep neural network architectures to tackle problems related to unstructured data such as images, audio, video, etc. Deep learning architectures such as autoencoders are popular in unsupervised image anomaly detection as they allow for the representation of data in forms where signal and noise can be separated into distinct layers by transforming the image into lower-dimensional embeddings. In this paper, we experiment with three different types of architectures assess their overall effectiveness in detecting anomalies in industrial images. 
\subsection{Motivation}
Our motivation for this project is inspired by three key pointers:
\begin{itemize}
     \item The sparsity of the data needed to learn about the anomalies is often a key challenge that is faced when it comes to deep learning methods since deep learning models rely on high volumes of data to extrapolate parameters that enable them to make strong generalizations about the information being fed into them. 
    \item Real-world data often comes in the unlabeled form, which is either encoded into a labeled format for machine interpretability or is left the way it is. This makes the task of training a deep neural network even more challenging and we were interested to investigate methods to tackle this challenge.
    \item Lastly, we wanted to choose a problem that is not just restricted to the academic realms of deep learning but is also applicable in the real-world setting such that the models we compare and contrast upon can be reflected well into business and industrial applications.
\end{itemize}
\section{Related Work}
\subsection{Supervised anomaly detection - CNNs}
Some of the key insights in the area of deep learning for anomaly detection spans from the basic conception of image processing neural networks that are highly utilized in the tasks of image recognition to adversarial learning methods that focus on game-theoretic architectures to generate and discriminate between images accurately. One of the most popular neural network architectures that are widely used today in deep learning is the convolutional neural network. Convolutional Neural Networks (CNN) is a special type of neural network created to work with two-dimensional image data. It can also be used with one- and three-dimensional image data. The convolutional layers perform an operation named “convolution”. It has many use cases, and some have already been discussed in the state-of-the-art papers \cite{stateoftheartanomaly}. One simple technique though is in computer vision and it can be used to detect scratches in images.
\par 
CNN Anomaly image detection is used in a lot of industries nowadays. For instance, the state-of-the-art papers address the main use cases of the CNN algorithms. It is used in finding faults, such as cracks, scratches, markings, missing parts, and inaccuracies in various inspection tasks \cite{stateoftheartanomaly}. Thus, involvement in factories and surface inspections are two key parts of a common application use case. One paper is drafting the issues and application of CNN in the automobile industry. A drastic growth of Automated Visual Inspection (AVI) systems in the industries has occurred in the past couple of years, with the aid of such applications to help human inspectors for anomaly localization and detection. Hence, the usage of CNN itself is not sufficient to solve the detection issues, since the anomaly datasets are usually relatively small and imbalanced \cite{salienttextural}.

Another common application of CNN is in the surveillance industry. The manual watching of toneless videotapes is not sufficient. What they need is an automated system. These alarm based systems detect anomalies by anomalous behaviors. However, knowing what an anomaly is and not is a difficult and complicated issue. Thus, it is necessary to set some boundaries and narrow down what is seen as an anomaly, like a car accident \cite{surveillancevideos}. Luckily, CNN has done this possible and a CNN-LSTM model achieves an accuracy of 85\%.The CNN architecture has existed since the 80s. However, it had a breakthrough in the 2000s with CPUs. Anyway, history has shown that adding more layers or making the model more complex does not necessarily improve the accuracy. Adding more layers is not necessarily helpful for improving the detection performance of a CNN model \cite{empiricalstudyanomaly}. They compared simple CNN models with different internal depths shallow CNN, moderate CNN, and deep CNN
\subsection{Unsupervised anomaly detection - Autoencoders}
Though CNN's are a powerful method of supervised anomaly detection, autoencoders have been shown promise in image anomaly detection due to their ability to represent a high dimension of features within low latent space and effectively reconstruct the image by focusing on maximizing the capture of useful information from the low dimension embeddings.
\paragraph{Reconstruction error}
Many papers mention the effectiveness of unsupervised approaches for anomaly detection in images, e.g. \cite{Beggel2020}, \cite{Bergmann2019}, \cite{Zenati2018}, \cite{Gong2019}. The core idea for unsupervised anomaly detection in images is to learn to create low-dimensional representations of an image, to reconstruct it and to compare it to the original image. When dimensionality reduction is learned using only normal images, the reconstruction error for abnormal images will likely be higher then for normal images. This is based on the assumption that anomalies can not easily be encoded in a latent representation that has been learned using only normal data points, thus cannot be reconstructed \cite{Zong2018}. Reconstruction error can therefore be used for detecting outlier images.
For this approach, many different implementations exist. One of the most conventional approaches for dimensionality reduction is principle component analysis (PCA), as described by \cite{Zong2018}. However, the most popular approach for anomaly detection through dimensionality reduction are autoencoders (AE), such as variational autoencoders \cite{Peng2020}, convolutional autoencoders \cite{Bergmann2019}, adversarial autoencoders \cite{Beggel2020} or autoencoders combined with other techniques, such as an autoencoding gaussian mixture model \cite{Zong2018} \par
Zhou and Paffenroth \cite{zhou2017anomaly} developed a robust deep autoencoder (RDA) that takes advantage of robust principal components analysis (RPCA) in order to develop a deep autoencoder that not only can discover high quality, non-linear features but is also good in eliminating outliers and noise without having any access to preprocessed training data. Baur et al. \cite{baur2018deep} developed a variational autoencoder (VAE) for the detection of anomalies in brain MRI scans and found that having constraints on the latent space and adversarial training can help to further improve the overall performance of the autoencoders, but also that the VAE offers no significant advantage against convolutional autoencoders. Zong et al. \cite{Zong2018} developed a deep autoencoding gaussian mixture model (DAGMM) for unsupervised anomaly detection which utilizes a deep autoencoder alongside a gaussian mixture model, and found that their model improves upon clDE). In KDE for anomaly detection data points that have a low probability density will typically be identified as outliers \cite{Peng2020}. However, it is hard to perform calculate the probability density function of high dimensional data, such as images. Dimensionality reduction can therefore be performed in advance, to obtain lower dimensional data, which can be used for KDE. Lv et al.\cite{Peng2020} describe that the combined use of a VAE and KDE yields promising results that outperform other approaches that use KDE with other dimensionality reduction methods.
\paragraph{Kernel density estimation} 
Another method for anomaly detection where autoencoders come into play, is kernel density estimation.
\subsection{Adversarial learning and anomaly detection : GAN}
Autoencoders are powerful due to their ability to compress the input down to a vector to a lower dimensionality and transform it back into a tensor with the similar shape in which it was fed into the network. But with advancements in generative learning models, one of the most prominent models out there that exists under generative learning are the family of generative adversarial networks. Developed by Goodfellow et al. \cite{goodfellow2014generative}, generative adversarial networks are a class of networks which are built around the concept of adversarial learning, a method of machine learning where models are trained by feeding them deceptive information in order to fool them and make them robust towards detecting ground truth and forged truth. With the advent of adversarial networks, the use of these models in anomaly detection has also been done quite extensively. Zenati et. al \cite{Zenati2018} developed a GAN-variant models for anomaly detection and found that their models were effectively able to achieve state-of-the-art performance on image and network intrusion datasets, while being several hundred-fold faster at test time than the only published GAN methodology. GAN's output, which are real-looking – but completely fake – data, can be especially useful as additional input data for datasets with small numbers of data. 
\section{Methodology}
\subsection{Data set}
The dataset that we used to train and test our model is the MVTec Anomaly Detection dataset provided by Bergmann \cite{Bergmann2019}. The dataset consists in images of objects and fabrics of 15 different classes, such as toothbrushes, screws, glass bottles and leather. For each class, a set of images of correct items for training and a set of images both correct and defective, labeled accordingly, for testing is provided. Depending on the class, there are between 50 and 300 images for training and a similar number for testing per class.
The images are RGB images with a resolution of 1024x1024 pixels. They have been taken in front of neutral backgrounds and often from the same angle and with the same detail for all objects. This is a very favorable condition for anomaly detection, as the difference in the pictures can be reduced to the anomalies that might be present. In a real life setting we would expect that there might be more differences between pictures. \par
In preparation for the model training we have reduced the size of the images, to either 256x256 or 128x128, we have experimented with different image sizes. Higher image size usually yields better results, however it slows down the training significantly and makes it more difficult to experiment with other parameters. 
In addition, images have been converted to grayscale for some models. 
We didn't augment images additionally by adding noise, scaling, squishing or rotating them for training purposes, however we tested the robustness of some models with noised images later in the process. More details on this can be found in the evaluation section.

\subsection{Evaluation metrics}
The images for testing provided in the MVTec AD dataset \cite{Bergmann2019} are labeled as good images and into different classes of defectious images. We summarize all defectious classes into one class and make the problem a binary decision problem: The task of all models will be to decide whether a presented image is an outlier or not. For evaluating the accuracy of the models on this task, we measure the F1-score. The F1-score is calculated using the following formula, where predicted outliers are positive and predicted normal images are negatives; therefore, an outlier, that has been detected will be a true positive (TP), an outlier that has not been detected a false negative (FN) and a normal image that was predicted as outlier a false positive (FP): $$F1-score=\frac{TP}{TP+\frac{1}{2}(FP+FN)}$$

\subsection{Architectures}
\subsubsection{Convolutional Neural Network}
Convolutional Neural Network (CNN) follows three key operations for the creation of the neural network architecture:
\begin{itemize}
    \item \textit{Convolutional Layers}: The convolutional layers focus around the usage of kernels, which are small in dimensionality and spread out along the overall depth of the input. In the convolution layers, a convolving operation takes place where each data point hits the convolutional layer, and the layer \textit{convolves} each filter across the spatial dimensionality of the input in order to re-create a two-dimensional activation map. With each gliding operation on the input data, the scalar product gets calculated for each value in the given kernel which allow the network to learn pixel-based characteristics about the network and determine what the image represents. Usually the rate at which the convolution takes place is decided by the stride parameter which denotes the number of dot products that need to be done.
    \item \textit{Pooling Layer}: The goal of the pooling layers is to further reduce the dimensionality of the image representation after it has passed through the convolutional layers. Within the pooling layers, the output of the convolutional layers is taken and depending upon the nature of the pooling layer, the pooling layer will operate over each activation map within the input and scale the dimensionality using the prescribed function.
    \item \textit{Fully-Connected Layer}: Lastly, the fully connected layer contains neurons which are directly connected to the output neurons, with their goal being to output the final predictions.
\end{itemize}
\subsubsection{Convolutional Autoencoders}
Autoencoders are a branch of deep learning networks where the outputs are an approximation of the input variables that are fed into the network. The inputs pass through a small junction of neurons within the hidden layers up to the bottleneck layer, usually called the encoder that goes to compress the input into a representation of the input variables. This compression is further unpacked and mapped to the output layers by the decoder that creates a sparse representation of the input that was provided. Autoencoders are useful in anomaly detection applications since they create representations that morph themselves close to the ground truth, and any deviation from this sparse representation leads to the detection of anomalies in images.  
\par 
Let us define the mathematical representation of an autoencoder. Let $N^{k}$ be the number of neurons in each layer $l$, $l = 1,2,...,L$. Let each output of neuron $n$ in layer $l$ as $\alpha_{n}^{(k)}$, with the vector of all the outputs for the given layer as $\alpha^{l} = (\alpha_{1}^{(l)},.....,\alpha_{N}^{(k)})$. For each hidden layer in the network, inputs from the preceding layer are compressed by the non-linear activation function $f(.)$ before being passed to the next layer. In order for the network initialization, the input layer focuses on the cross sections of the previous layer's output as raw predictions, $\alpha^{(0)} = \alpha = (\alpha_{1},\alpha_{2},...,\alpha_{t})$. Because this process keeps on repeating itself in a recursive fashion, the output formula for the neural network in layer $l > 0 $ will be defined as:
\begin{equation}
    \alpha^{(l)} = f(b^{(l-1)} + W^{(l-1)}\alpha^{(l-1)})
\end{equation}
Where $W^{(l-1)}$ is represented as a $N^{(l)} \times N^{(l-1)}$ matrix of weight parameters, and $b^{(l-1)}$ is the bias vector of dimension $K^{(l)} \times 1$. With the compression done, the final output from the autoencoder is defined as: 
\begin{equation}
    F(\alpha, b, W) = b^{(L)} + W^{(L)}\alpha^{(L)}
\end{equation}For our project, we focused on using the rectified linear unit activation function (ReLU) since it is the most commonly used activation function and forces all outputs to be strictly positive, the function being mathematically defined as the maximum of the target output and 0, $max(y, 0)$.
\begin{figure}
  \includegraphics[width=4cm]{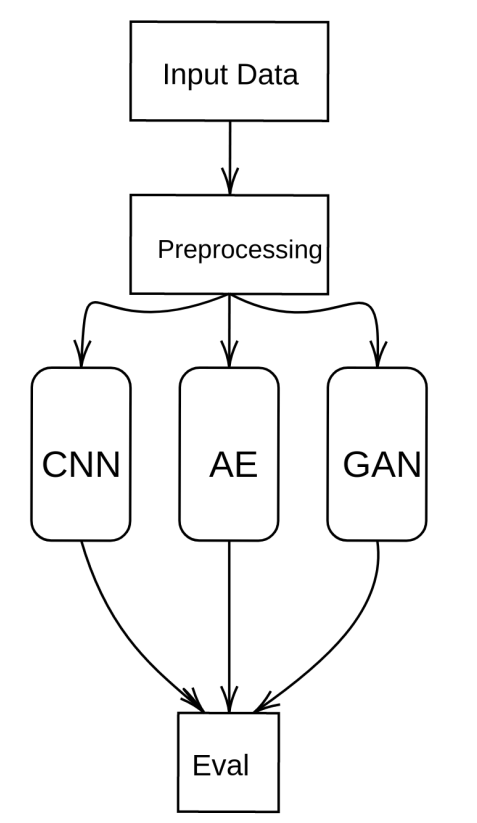}
  \caption{Layout for the assessment of anomalies in MVTec Dataset.}
  \label{fig:teaser}
\end{figure}

\subsubsection{Generative Adversarial Networks}
Generative Adversarial Networks (GANs) are a technique for both semi-supervised and unsupervised learning, and they achieve good results by implicitly modelling high-dimensional distributions of data. GANs can be characterized by training a pair of networks competing with each other, in which one could imagine that one network is an art forger and the other network is an art expert. The forger, which is the generator in GANs, creates forgeries with the aim of making realistic images, and the inspector, which is named as discriminator, aims to tell the fakes apart from all images. Ultimately, the last rendition of forgeries that fooled the inspector is displayed at the gallery, aka the GAN's output. 
\par
The generator network model we have here is: $x=G(z; \theta^{G})$, where $z$ is the fake data, $G$ is the function model that can make $z$ into a "real" data $x$. In the training procedure, we can use any SGD-like (Stochastic Gradient Descent) algorithm on two minibatches simultaneously, one is the genuine data set and the other one generated data set. In real training, we ran $k$ steps of one player for every step of the other player, in that it can prevent one player from falling behind too much. The objective function to judge the performance of the model is: 
\begin{equation}
J^{(D)}=-\frac{1}{2}\times E_{x~p_{data}}\log D(x)-\frac{1}{2}E_{z}\times \log(1-D(G(z)))
\end{equation}
\begin{equation}
J^{(G)}=-J^{(D)}
\end{equation}
\par Here $J^{(D)}$ is the discriminator's objective function, a cross entropy function, and the left represents that $D$ successfully determined $x$ is a genuine $x$, and the right part represents that $D$ thinks $z$ is genuine. Similarly, $J^{(G)}$ is the objection function of the generative network, whose aim is opposite to that of $D$, thus there is a $minus$ in front. Here we want to maximize $D(x)$ for every image from the true data distribution $x~p_{data}$, as well as minimize $D(x)$ for every image not from the true data distribution $x-p_{data}$. 

\par Actually this process is a manimax game or zero-sum game, in which one needs the largest and one needs the smallest, and we want to find the equilibrium point, which is the saddle point of $J^{(D)}$.

\begin{figure}[H]
    \minipage{0.50\textwidth}
        \centering
        \includegraphics[height=3cm]{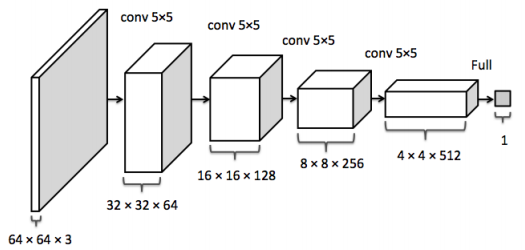}
        \caption{The discriminator convolutional network}
        \label{toothbrush}
    \endminipage\hfill
\end{figure}

\par A clearer function to show the process can be shown as: the optimal $D(x)$ for any $P_{data}(x)$ and $P_{model}(x)$ is always:
\begin{equation}
D(x) = \frac{P_{data}(x)}{P_{data}(x)+P_{model}(x)}
\end{equation}
\par And we want the value of $D(x)$ to be infinitely near to $\frac{1}{2}$, which means the $P_{data}$ is infinitely near to $P_{model}$, which is what we want and which is when $D(x)$ cannot discern the fake from the genuine.
\par 
The GANs model we use here is Deep Convolutional GANs (DCGANs), which allows training a pair of deep convolutional generator and discriminator networks. The objective function is similar to the previous one but the $J^{(G)}$ is different:
\begin{equation}
J^{(G)} = -\frac{1}{2}\times E_{z}\log D(G(z)) 
\end{equation}
\par Here it's called a Non-Saturating Game, in which we use the fake data success rate of $G$ itself as it's own objective function instead of the minus of $J^{(D)}$. In this way, the equilibrium is not decided by loss anymore and there is no unnecessary bond between $J^{(D)}$and $J^{(G)}$ anymore, which mean $G$ can still be optimized once $D$ is in its optimal scenario. Through SGD algorithm, the optimization is realized by the following 3 main steps:
\begin{itemize}
    \item Keep $G$ still, train $D(x)$ so that the function $(3)$ is maximized
    \item Keep $D$ still, train $G$ so that the function $(6)$ is maximized
    \item Repeat step $1$ and $2$ until $D$ and $G$ reached equilibrium
\end{itemize}

DCGANs make use of strided and fractionally-strided convolutions which allow the spatial down-sampling and up-sampling operators to be learned during training. These operators handle the change in sampling rates and locations, which is a key requirement in mapping from image space to possibly lower dimensional latent space, and from image space to a discriminator.

The process behind a Deep Convolutional GANs includes:
\begin{itemize}
    \item Replace any pooling layers with strided convolutions (discriminator) and fractional-strided
    \item Use batchnorm in both the generator and the discriminator
    \item Remove fully connected hidden layers for deeper architectures
    \item Use ReLU activation in generator for all layers except for the output, which uses the hyperbolic tangent activation function. 
    \item Use LeakyReLU activation in the discriminator for all layers
\end{itemize}

The ultimate goal for our DCGAN was to generate more input data. This is because some datasets, like the Toothbrush dataset, only had 60 images in it.

\subsection{Experimentation: Supervised Methods}
\subsubsection{Preprocessing} For the CNN we will use the \textit{ImageDataGenerator} to make the training and validation data. Additionally, it will also help with data normalization.
\subsubsection{Five convolution layers} The convolutional layer is the core building block of a CNN. The layer's parameters consist of a set of learnable filters (or kernels), which have a small receptive field, but extend through the full depth of the input volume. In our case, we will stack it with one convolutional layer, with input shape 200x200x3, kernel of 3x3, 16 filters, and the activation function ReLU (Rectified Linear Unit), followed by a max pooling layer, with input shape 100x100x3. This will be repeated 5 times.
\subsubsection{One flatten layer} Reshapes the tensors and flattens the input. Thus, we will have one flattens layer that flattens the output of the CNN convolution layers.
\subsubsection{One hidden layer} Hidden layer with an activation function to be sent to the output layer.
\subsubsection{One output layer} Converts the result to a binary result. Either it is true or false based on the sigmoid activation function.
\subsubsection{Model specifications}: As normally combined with CNN, we will use the RMSprop with Binary Crossentropy Loss to train our model. A good property of the RMSprop is that it is automating the tuning of the learning rate.

\subsection{Experimentation: Unsupervised methods}
\subsubsection{Reconstruction error and kernel density using convolutional autoencoder}
For this part, we propose a variation of the autoencoder named as KD-CAE, a convolutional autoencoder that utilizes kernel density estimation alongside reconstruction error. This approach was inspired by the work of Lv et al. \cite{Peng2020} who highlight the potential of kernel density estimation for anomaly detection. The model that we developed consists of 12 layers, 6 in the encoder and 6 in the decoder. The layers are alternating between convolutional layers, with increasing convolutions starting at 32 convolutions and going up to 128 convolutions in the last layer of the encoder. The dimensions of the images are reduced from 128x128 to 8x8. This means, that the flattened representation of the latent space is a vector of length 8192. This seems to be quite big for a latent representation, however, reducing the dimensions even more led to less accurate image reconstructions for all images, normal and anomalies, thus the performance of anomaly detection was worse.

\subsubsection{Noise Injected Convolutional Autoencoder}
In order to assess the robustness of our autoencoder models, our next step was to determine if the autoencoders are robust to noise that can come through in the images. Hence, in order to mimic real life discrepancies in image clarity, we modified the convolutional autoencoder with a noise injection function, called NI-CAE. We injected Gaussian noise into the images, with a mean parameter of 0 and variance parameter of 0.001. To make the noise injection be more reflective of real world conditions, random images would be selected for the injection of noise, the random selection being decided by the following formula:
\begin{equation}
    S_{r} = Random(K_{D}, [0.10 * K_{D}]) 
\end{equation}
Where $S_{r}$ are the sample indices that are randomly picked from the image count for a given image type, with the number of image indices sampled by calculating the 10\% proportion value of the overall size of the sample for the given image.
\par 
The architecture of the NI-CAE is specified to have an encoder of five Conv2D with a convolution filter size of $3\times3$, the layers descending in neuron size as [128,64,16,8,4], five MaxPooling2D layers of a $2\times 2$ filter, squeezed through a Dense layer of size 512. For the decoder, the exact same architecture as the encoder is used but is flipped the other way round, with the final output layer having a convolution of $1\times1$. In order to prevent overfitting, Early stopping was kept to stop the model from training if the validation score does not improve over 5 epochs. No regularization layers were added to the autoencoder. Also, the images were grayscaled and normalized in order to make them all uniform.
\section{Evaluation}
\subsection{CNN results}
We tried the CNN on a couple of datasets as well. However, we did not expect it to perform well, given all the circumstances around how the algorithm performs. However, we got some results, and the Toothbrush dataset and got results of ROC AUC score around 65\% in best cases and an F1 score around 25\%. Anyway, the results confirm that the model alone is not sufficient and has difficulties achieving what we are seeking in this paper.\\
\begin{table*}[h]
\begin{tabular}{|l|l|l|l|l|l|}
\hline
         & Toothbrush & Bottle & Screw & Leather & Transistor \\ \hline
ROC AUC  & 65\%       & 31\%   & 29\%  & 2\%     & 70\%       \\ \hline
F1 score & 25\%       & 9\%    & 7\%   & 7\%     & 4\%        \\ \hline
\end{tabular}
\caption{Results of the CNN}\label{table:cnn}
\end{table*}

% \begin{figure*}[h]
%     \centering
%     \includegraphics[width=0.5\textwidth]{CNN_Toothbrush_ROC_curve.png}
%     \caption{CNN ROC Curve of the Toothbrush dataset}
%     \label{fig:CNN_Toothbrush_ROC_curve}
% \end{figure*}

\subsection{KD-CAE}
The convolutional autoencoder with which we used reconstruction error and kernel density estimation as indicators for detecting anomalous images yields very different results, depending on the class that it was tested on. Table \ref{table:kd_cae} summarizes all the results. 
\begin{table*}[h]
\centering
\begin{tabular}{|l|l|l|l|l|l|}
\hline
                                & Toothbrush & Bottle & Screw & Leather & Transistor \\ \hline
Reconstruction error threshold  & 0.005      & 0.004  & 0.004 & 0.003   & 0.0055     \\ \hline
Kernel density threshold        & 5630       & 5600   & 5625  & 5651    & 5350       \\ \hline
Train size                      & 60         & 167    & 256   & 254     & 171        \\ \hline
Test size                       & 42         & 86     & 160   & 124     & 100        \\ \hline
Proportion of anomalies in test & 71\%       & 77\%   & 75\%  & 74\%    & 40\%       \\ \hline
F1                              & 0.85       & 0.87   & 0.11  & 0.72    & 0.59       \\ \hline
\end{tabular}
\caption{Results of the KD-CAE}\label{table:kd_cae}
\end{table*}

\begin{table*}[]
\begin{tabular}{|l|l|l|l|l|l|}
\hline
                           & Toothbrush & Bottle & Screw & Leather & Transistor \\ \hline
F1(Without Noise) & 0.7347              & 0.3703          & 0.1221         & 0.4203           & 0.8083              \\ \hline
F1 (With Noise)   & 0.5282              & 0.2412          & 0.083          & 0.3394           & 0.7876              \\ \hline
\end{tabular}
\caption{Results of the NI-CAE}
\end{table*}

\begin{figure*}[h]
    \centering
    \includegraphics[width=0.4\textwidth]{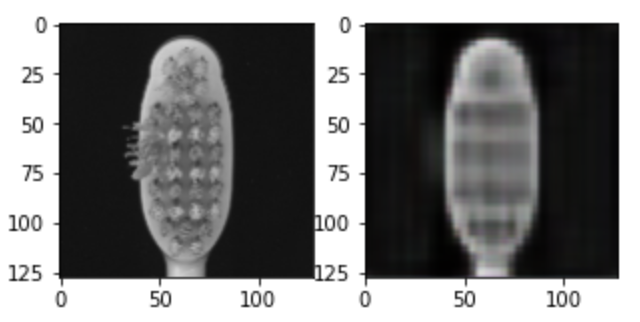}
    \includegraphics[width=0.4\textwidth]{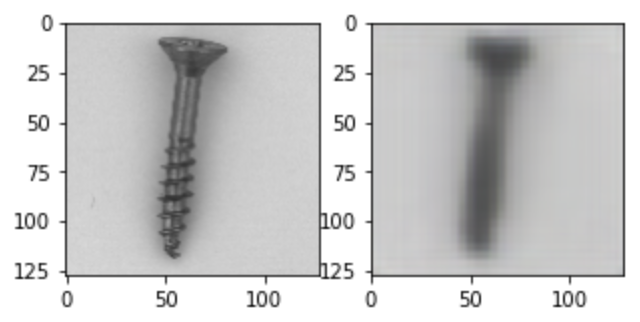}
    \includegraphics[width=0.4\textwidth]{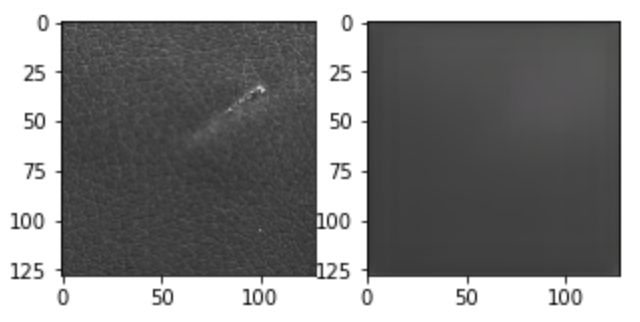}
    \includegraphics[width=0.4\textwidth]{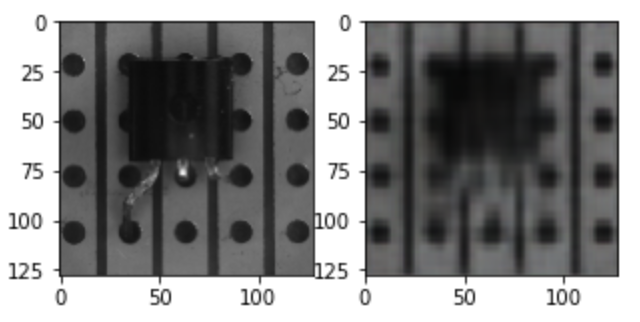}
    \caption{Examples of original and reconstructed anomalous images}
    \label{fig:reconstruction_examples}
\end{figure*}

\begin{figure*}[h]
    \centering
    \includegraphics[width=0.5\textwidth]{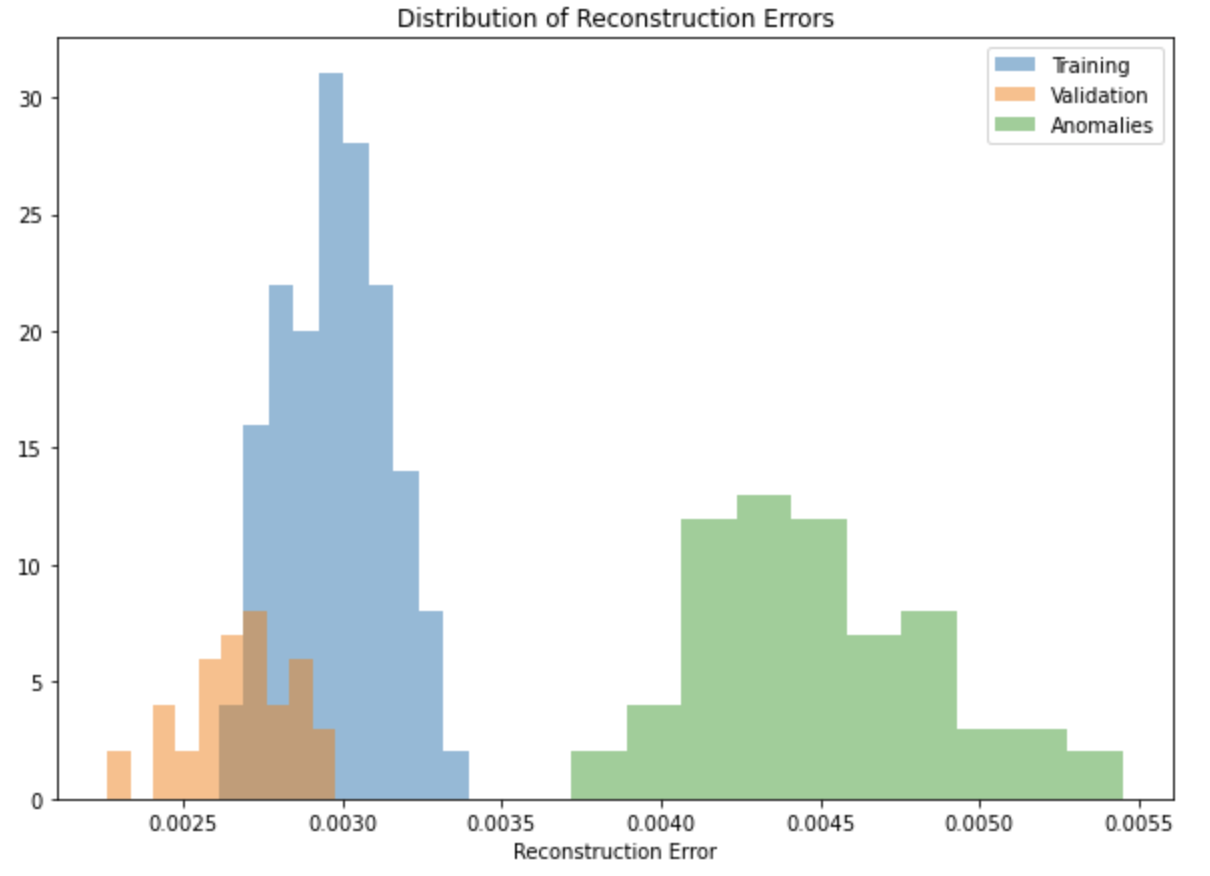}
    \caption{Reconstruction error distribution on the bottle dataset using the KD-CAE}
    \label{fig:reconstruction_errors_bottle}
\end{figure*}

Figure \ref{fig:reconstruction_examples} shows examples of original images and reconstructed images using the convolutional autoencoder. Even though there is quite an information loss due to the encoding and decoding process, we can see that the autoencoder removes things that weren't expected as for example the protruding bristles in the toothbrush or the scratch in the leather. Furthermore, the autoencoder makes the reconstructed images look like they were created from normal data points. We take advantage of this behavior by calculating the reconstruction error, or in other words, calculating the difference between the original image and the reconstructed image. For objects where our autoencoder has well performed, the reconstruction error is consistently higher on anomalous images than on training and validation data. The reconstruction error on bottles is illustrated in figure \ref{fig:reconstruction_errors_bottle} as an example.\\
The autoencoder is having problems to reconstruct objects that are not fixed in the image, as for example the screws, which are in a different position in every image. Additionally, the errors in the screws are very small, they are in fact barely visible to the human eye. Unfortunately, the error that is made only by reconstructing screws, anomalous or not, dominates the overall error, as anomalies in screws are so small. It is, therefore, only logical that the F1-score for anomaly detection is very low.
Given the fact, that we have only started experimenting with the autoencoder and we had strong computational constraints, results are quite promising. The model needs to be improved for detecting very small damages and being more robust for objects being in different positions in the image. But it can be concluded that it shows strong performance if certain constraints are fulfilled, such that the object is always in the same position and the defect on the object is not too small.

\subsection{NI-CAE}
For the noise injected convolutional autoencoder, what we found is that the injection of noise into the training sets drastically impacts the performance rate of the CAE, as established in Table 3 . With the addition of Gaussian noise, the accuracies for all the image types goes down, with the greatest drop seen in the Toothbrush dataset, a total of 28.10\% reduction in the F1-score. The drop in F1-score for bottle and and leather was seen to be normal, but the transistor dataset turned out to be the most robust to noise injection, only showing a reduction of 2.56\% from noiseless images to noise induced images. Though compared to KD-CAE, NI-CAE performed much better on the transistor dataset but failed to outperform for the rest of the image sets. Nonetheless, the comparison aids us to determine that the inclusion of the noise does infact change the way the autoencoder learns the representations and that itself can change how the network learns to reconstruct the images. Figure 5 helps to visualize how addition of noise in the dataset changes the reconstructed output, comparing it against the ground truth and the SSIM difference which was computed using the SSIM scoring \cite{wang2004image}. 

\begin{figure*}[h]
    \centering
    \includegraphics[width=\textwidth]{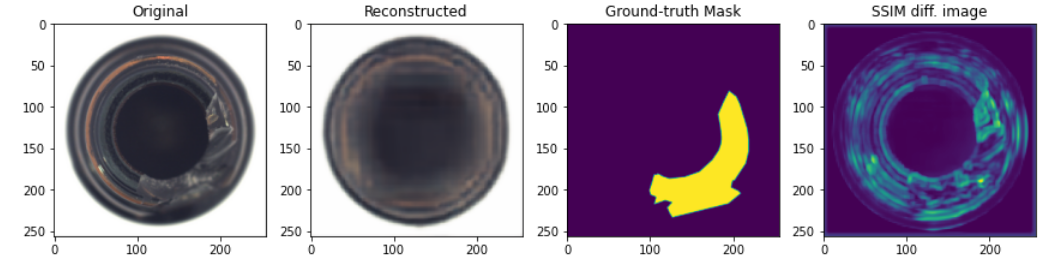}
    \caption{Image reconstruction of the bottle dataset, alongside the ground truth mask and the SSIM difference image}
    \label{fig:my_label}
\end{figure*}

\subsection{GAN}
The results of our DCGAN can be seen in the Figure \ref{fig:toothbrush_GAN} below, which depicts toothbrushes created by our model. Note that these are generated on 60 images of toothbrushes, yet look unmistakably real to the naked human eye. Adding more real-looking images, or real-looking images with anomalies, to models described above can make the results more robust.    

\begin{figure*}[h]
    \centering
    \includegraphics[width=0.5\textwidth]{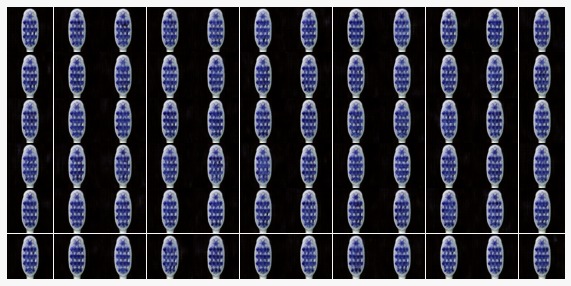}
    \caption{Toothbrush images generated by GAN}
    \label{fig:toothbrush_GAN}
\end{figure*}

\section{Conclusion}
We compared a supervised method with non-supervised methods for anomaly detection. A CNN was used as the supervised method. The model was trained on anomalous and normal labeled data, similar to a normal classification problem. The disadvantage is that this method requires a lot of examples of anomalous data, which are not necessarily available.
Therefore, we decided to tackle the problem of anomaly detection head-on in an unlabeled setting in order to truly mimic how real-world intelligent systems are supposed to behave in intrusion detection settings. And since our focus is on the detection of anomalies in images, we chose to utilize autoencoders in order to perform unsupervised learning and create a latent representation of our target images so that we can quickly identify anomalous regions in our images without worrying about the issue of image labeling. 
We could show that the CNN gives quite poor results, which can be explained by the fact that it lacks data. The autoencoders, however, show promising results, which confirms our hypothesis that unsupervised anomaly detection would be a good option if sufficient anomalous data is lacking.
An additional idea on how to tackle the problem of data limitation in supervised settings was to utilize the state of the art general adversarial network (GAN) as a medium of creating new batches of image samples that can be used in conjunction with anomaly detection. We were able to train a GAN that produces new images, however didn't have the capacities left to extensively test these images for training in a supervised setting. This is a potential idea for future work on this topic. 

\paragraph{A word on overfitting}
Overfitting takes a very special role in the frame of this project. Usually you try to prevent overfitting from happening, in order to be able to generalize to new data, that might be different from the data that was seen during the training process. In this case, we want to avoid to generalize too much to new data, because this would mean that the model is able to reconstruct anomalous data just as good as data similar to the data seen during the training, thus detecting the anomalous data would in fact become more difficult. Therefore, a little bit of overfitting is even benefitting the purpose of anomaly detection.
\bibliographystyle{ACM-Reference-Format}
\bibliography{sample-base}

\appendix

\end{document}